# Improving Indonesian Text Classification Using Multilingual Language Model

Ilham Firdausi Putra
School of Electrical Engineering and Informatics
Institut Teknologi Bandung
Indonesia
ilhamfputra31@gmail.com

Ayu Purwarianti[1,2]
[1]U-CoE AI-VLB
[2]School of Electrical Engineering and Informatics
Institut Teknologi Bandung
Indonesia
ayu@stei.itb.ac.id

*Abstract*— Compared to English, the amount of labeled data for Indonesian text classification tasks is very small. Recently developed multilingual language models have shown its ability to create multilingual representations effectively. This paper investigates the effect of combining English and Indonesian data on building Indonesian text classification (e.g., sentiment analysis and hate speech) using multilingual language models. Using the feature-based approach, we observe its performance on various data sizes and total added English data. The experiment showed that the addition of English data, especially if the amount of Indonesian data is small, improves performance. Using the fine-tuning approach, we further showed its effectiveness in utilizing the English language to build Indonesian text classification models.

*Keywords—multilingual language model, text classification, sentiment analysis, hate speech classification, Indonesian text*

## I. INTRODUCTION

With the development of Indonesian people's access to the internet, more and more text data is available digitally. This data is full of information and very useful if processed. For example, for business owners, citizen comments on the internet can be analyzed for sentiment to determine their reaction to something. Then for those who have a website, detecting violations in online conversations such as hate speech or abuse automatically can be very helpful.

Building such good automated text classifiers needs a lot of labeled data, which the Indonesian language lacks compared to other languages such as English. For example, the tasks of sentiment analysis and hate speech detection have English dataset like Yelp Review[1] (598K) and Jigsaw Toxic Comment[2] (1902K) respectively. In contrast with the Indonesian language dataset from recent publications such as [2] (12K) & [3] (11K) for sentiment analysis, and [4] (12K) for hate speech detection.

Recent deep, contextualized language models trained on corpora of multiple languages provide powerful, general-purpose linguistic representation across language [1, 5]. These language models were trained on a corpus of multiple languages without any multilingual objective. Surprisingly, an empirical investigation has shown that its hidden representation does share a common subspace that represents useful linguistic information in a language-agnostic way [6]. Based on these findings, we seek to improve the Indonesian language's lack of data with English data by utilizing multilingual language models.

In this paper, we investigate the effect of combining English and Indonesian data on building Indonesian sentiment analysis and hate speech detection using multilingual language models. Our findings show that the addition of the English dataset using the feature-based approach, especially if Indonesian language data is small, improves Indonesian text classification performance. However, there are cases where the addition of excessive English data decreases classification performance. Using a fine-tuning approach, we further improve the result of previous research on sentiment analysis [2, 3] and hate speech detection [4] on the Indonesian language.

## II. RELATED WORKS

Previous works have conducted Indonesian sentiment analysis using various text representation and model approaches. Farhan and Khodra [2] conducted experiments on various text representations with support vector machine, random forest, and multilayer perceptron as a model. Evaluation of their dataset yields 0.8521 F1-score using a neural network with TF-IDF as its feature. Crisdayanti and Purwarianti [3] conducted experiments on various text representations and neural network topology models. Evaluation of their dataset yields 0.9369 F1-score using Bi-LSTM with word embedding enhanced with paragraph vector. Ibrohim and Budi [4] conducted experiments on multi-label hate speech and abusive language in Indonesian twitter. Evaluation of their dataset yields 77.36% average accuracy using random forest with unigram as its feature. Previous works on text classification using multilingual language models have focused on its zero-shot capability across 15 XNLI languages [5]. Using its largest multilingual model, [5] showed that their model is very competitive and even outperforming the monolingual language model in various languages. There is no research in Indonesian text classification using multilingual language models, especially in sentiment analysis and hate speech classification.

## III. MULTILINGUAL LANGUAGE MODEL FOR INDONESIAN TEXT CLASSIFICATION

The latest Indonesian text classification model has been developed using representation from word embeddings and sequential models such as recurrent neural networks (RNN), long short-term memory (LSTM) [13] or, gated recurrent unit (GRU) [14] neural networks. While the combination of word embedding with RNN, LSTM, or GRU as the model has been very successful, it still possesses some limitations [1, 8]. First, the nature of sequential processing restricts the model since it can only attend to the previous token. Second, pre-trained word representation failed to capture deeper meaning on a document of text. Both problems were solved using the recently developed language model, such as BERT [1].

---

[1] https://www.yelp.com/dataset

[2] https://www.kaggle.com/c/jigsaw-unintended-bias-in-toxicity-classification/data

BERT use Transformer architecture and pre-trained on millions of texts with masked language model (MLM) objective. MLM randomly masked a token on its input and asked the model to predict the masked token. This novel objective, paired with the Transformer and its self-attention mechanism, allows the model to learn from the context of the whole document. This method has proven massively successful, topping various benchmark and significantly improve the performance from previously state-of-the-art sequential based language model such as ELMo [15].

On the other hand, labeled Indonesian text data is scarce in comparison to English text data. Unfortunately, cross-lingual representation of text has enabled models to do transfer learning across languages. It is now possible to train the model on one language, and fine-tune it to a downstream task in another language. To utilize the English language data on the Indonesian language task, we need a cross-lingual representation of English and Indonesian language on shared vector space.

There are two main methods of producing cross-lingual representation on shared vector space: *alignment* and *joint optimization* [9]. Earlier works on producing cross-lingual representation have relied on the former. Mikolov, Le, & Sutskever [10] trained an embedding of two different languages independently and aligned the two using small bilingual data. While it had great success, [9] argues that there are critical downsides such as it rely on having good parallel data and a key assumption that the embedding spaces of each language are isomorphic, which [11, 12] proves does not hold for many language pairs

The recent development on joint training methods has shown great results in obtaining cross-lingual representation without any parallel data. The method simply trains Transformer based language models on corpora of multiple languages, without any multilingual objective. The resulting models have shown good generalization ability across language and language-agnostic representation inside its hidden state [6]. Using this cross-lingual and strong pre-trained representation, we add English text data to train our Indonesian text classification model.

One such multilingual language model that we used in the experiments is Multilingual BERT [1] and XLM-RoBERTa [5] (henceforth, mBERT and XLM-R). Denoting the number of Transformer blocks as L, the hidden size as H, and the number of self-attention heads as A, Multilingual BERT was using the BERT-base architecture (L=12, H=768, A=12, Total Parameters=110M) and was trained on a concatenation of Wikipedia corpora from 104 languages. Because the size of Wikipedia for a given language may vary greatly, a certain language like English may be over-represented. To counteract the imbalance, exponentially smoothed weighting of the data was performed to over-sample low recourse languages and under-sample high-resource languages.

The XLM-R model was designed to improve mBERT in various ways. It follows the improvement in the BERT pre-training approach [7]. It was trained with dynamic masking, longer sequences, bigger batches, and without the Next Sentence Prediction objective. The XLM-R Large also has substantially more parameters and was trained on a larger balanced dataset from the CommonCrawl corpus that contains 100 languages. The XLM-R variant that we use in the experiment is XLM-R Large (L = 24, H = 1024, A = 16, 550M params).

IV. EXPERIMENTS

In this section, we describe the building blocks of our experiment. Fig. 1. show the overview of the experiment. Overall, it consists of three training data scenarios, two training approaches, and five datasets.

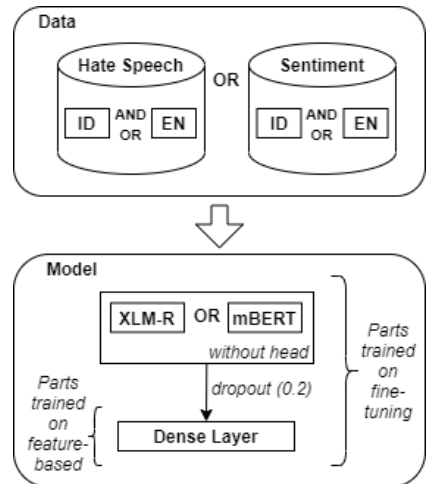

Fig. 1. Experiment overview

*A. Training Data Scenarios*

We investigate the model performance in three different scenarios. Each differs by the combination of the language used in its training data: monolingual, zero-shot, and multilingual. In the monolingual scenario, we use the Indonesian language text to train and validate the model. In the zero-shot scenario, we use the English language text to train the model while being validated on Indonesian text. Lastly, we use a combination of Indonesian and English text to train the model while being validated on Indonesian text in the multilingual scenario. Using these scenarios, we observe the improvement of the added English text.

*B. Training Approaches*

There are two approaches on applying large pre-trained language representation to downstream tasks: *feature-based* and *fine-tuning* [1]. On the feature-based approach, we extract fixed features from the pre-trained model. In this experiment, we use the last hidden state, which is 768 for mBERT and 1024 for XLM-R Large, as the feature. This extracted feature is then fed into a single dense layer, the only layer we trained on the feature-based approach, connected with dropout before finally ending on a sigmoid function. In contrast, the fine-tuning approach trains all the language model parameters, 110M for mBERT and 550M for XLM-R Large, including the last dense layer, on the training data binary cross-entropy loss.

Using the feature-based scenario, we run many experiments as the expensive and multilingual representation have been precomputed on all the data. In all training data scenarios, we vary the total data used. More specifically, we train the model using [500, 1000, 2500, 5000, 7500, Max] text data. Specific to multilingual training data scenario, we vary the amount of added English data by [0.25, 0.5, 0.75, 1, 1.5, 2, 3, 4, 5, 6, 7, 8, 9, 10] times the amount of Indonesian text data. We refer to a multilingual experiment with added English data N times the amount of Indonesian text data as multilingual(N).

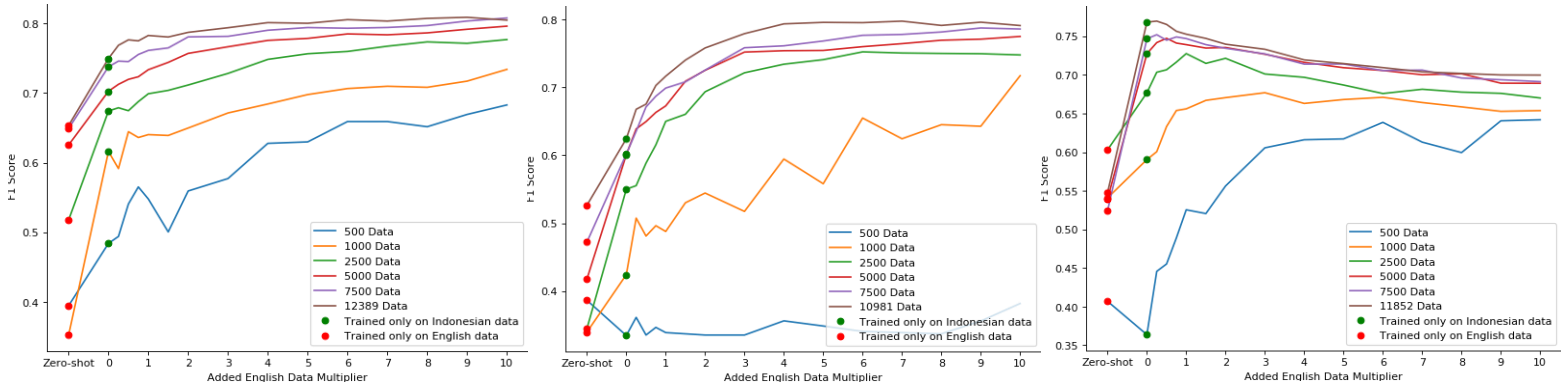

Fig. 2. Feature-based experiment result with XLM-R on [2] (left), [3] (middle), and [4] (right)

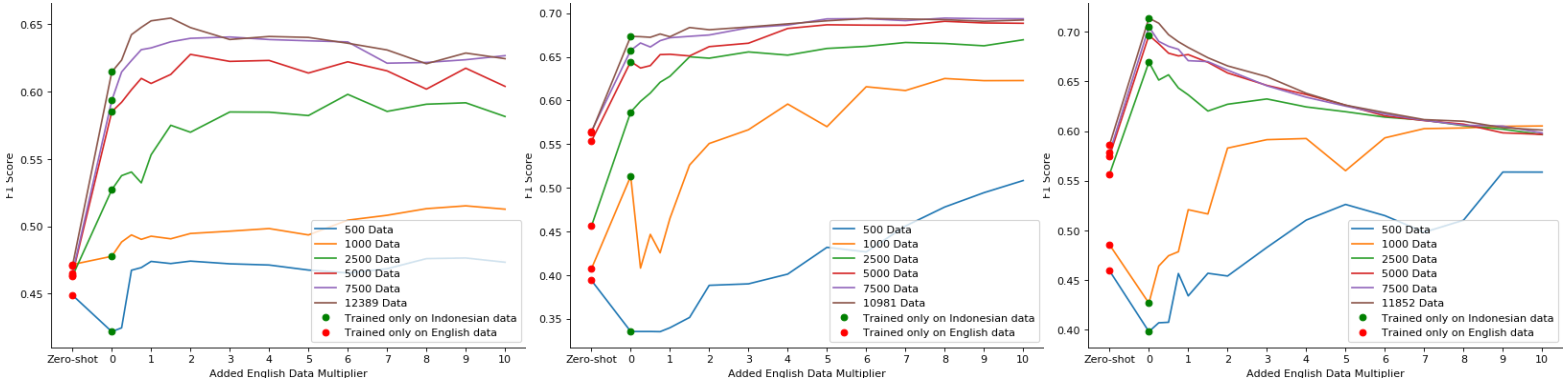

Fig. 3. Feature-based experiment result with mBERT on [2] (left), [3] (middle), and [4] (right)

In contrast to the feature-based scenarios, fine-tuning the full language model is expensive and resource-intensive. However, as shown in [1], fully fine-tuning the full language model will result in a better text classifier. We fine-tuned the best performing model on the feature-based scenarios. The experiment was reduced to only using the maximum total data and an added English data multiplier up to 3.

### C. Datasets

TABLE I. SENTIMENT DATASET DETAILS

| Data | | Train | Test |
|---|---|---|---|
| Source | Label | | |
| Farhan & Khodra [2] | Positive | 6281 | 1125 |
| | Negative | 6108 | 1304 |
| Crisdayanti & Purwarianti [3] | Positive | 7151 | 208 |
| | Negative | 3830 | 204 |
| Yelp Review | Positive | 299000 | N/A |
| | Negative | 299000 | N/A |

For the Indonesian language sentiment analysis dataset, we used data originated from previous related work by [2, 3]. Farhan and Khodra [2] used reviews crawled from TripAdvisor. Crisdayanti and Purwarianti [2] used text from Twitter, Zomato, TripAdvisor, Facebook, Instagram, and Qraved. For the English language dataset, we used data from Yelp that they have open-sourced. The details on the train, test, and label distribution for each data can be seen in Table I.

For the Indonesian language hate speech dataset, we used data originated from previous related work by [4]. Ibrohim and Budi [4] crawled tweets from Twitter and annotated the text with the help of 30 diverse annotators. The annotation not only includes whether or not the text falls into hate speech, abusive, or normal, but also its target, category, and level. For the English language, we use data from Jigsaw. The data comes from various conversations over the internet, annotated by dozens and up to thousands of annotators per text. The data also has rich and detailed labels up to its level and category.

TABLE II. HATE SPEECH DATASET DETAILS

| Data | | Train | Test |
|---|---|---|---|
| Source | Label | | |
| Ibrohim & Budi [4] | Hate speech | 6578 | 731 |
| | Normal | 5274 | 1586 |
| Jigsaw Toxic Comment | Hate speech | 152111 | N/A |
| | Normal | 1750083 | N/A |

However, the fine-grained labels between the English and the Indonesian hate speech data did not exactly match. More importantly, the two datasets differ in their main label. Ibrohim and Budi [4] annotate the text into normal, abusive, or hate speech. Meanwhile, Jigsaw annotates the text into normal or toxic. We simplify the label of [4] into a simple binary label indicating the text is normal or hate speech/abusive. The hate speech dataset details can be seen in Table II.

*D. Training Reproducibility and Hyperparameters*

From the dataset detailed in Table I & II, the training data further split into training and validation set with a 90:10 ratio. The split was done in a stratified fashion, conserving the distribution of labels between the training & validation set. The result is a dataset separated into training, validation, and test sets.

Each experiment will train the model using the training set and validate it to the validation set on each epoch. After each epoch, we will evaluate whether we will continue, reduce the learning rate, or stop the training process based on validation set performance and the hyperparameter set on each condition. In the end, we use the model from the best performing epoch based on its validation performance to predict the test set.

On the feature-based experiment, we set the final layer dropout probability to 0.2, the learning rate reducer patience to 5, and the early stopping patience to 12. On full fine-tune experiment, we set the final layer dropout probability to 0.2, the learning rate reducer patience to 0, and the early stopping patience to 4. Every validation and prediction use 0.5 as its label threshold.

To ensure reproducibility, we set every random seed possible on each experiment. On the feature-based experiment, we average the result of 6 different runs by varying the seed from 1-6. Running the same experiment on the feature-based approach will result in the same final score. On the full fine-tune experiment, we only run one experiment. While the result should not differ substantially, the exact reproducibility cannot be guaranteed as the training was done on a TPU[3].

## V. RESULTS

We build the system as mentioned in the previous section. The next subsection talk in detail about the result of each training approach.

*A. Feature-based experiment*

The result of the feature-based experiment with XLM-R model on all datasets can be seen in Fig. 1. Through this result, we can see that adding English data can help the performance of the model. On [2] & [3] dataset, adding English language data consistently improves the performance. However, on [4], there is a point where the added English data results in worse performance. We hypothesize this is due to the large difference in what constitutes hate speech (or toxic by Jigsaw dataset) between the datasets used.

The result of the feature-based experiment with mBERT model on all datasets can be seen in Fig. 2. The same phenomenon is observed on mBERT based experiment, although the performance is substantially lower. This is expected as XLM-R is designed to improve mBERT on various design choices.

Defining the gain as the difference between monolingual and its highest multilingual performance, Table III shows the gains averaged on all datasets across total data and model. The highest gain can be seen on the lowest amount of total data used, 500, with an F1-score gain of 0.176 using XLM-R model and 0.129 using mBERT model. The results suggest that the lower the amount of data used; the more gains yield by adding English data to the training set.

TABLE III. AVERAGE F1-SCORE GAINS

| Total Data | Average Gain | |
|---|---|---|
| | *XLM-R* | *mBERT* |
| 500 | 0.176221 | 0.129394 |
| 1000 | 0.165718 | 0.109215 |
| 2500 | 0.118456 | 0.051226 |
| 5000 | 0.095780 | 0.029564 |
| 7500 | 0.086930 | 0.028043 |
| MAX | 0.077875 | 0.020184 |

*B. Full fine-tune experiment*

The result of fully fine-tuning all parameters, in addition to utilizing English data, proved to be effective in building a better Indonesian text classification model. On [2] dataset, the highest performance achieved on the zero-shot scenario where it yielded 0.893 F1-score, improving the previous works of 0.834. On [3] dataset, the highest performance achieved on multilingual(1.5) scenario where it yielded perfect F1-score, improving the previous works of 0,9369. On [4] dataset, the highest performance achieved on multilingual(3) scenario where it yielded 0.898 F1-score and 89.9% accuracy. To provide a fair comparison with the previous work by Ibrohim & Budi [4], we also ran the experiment using the original label and monolingual scenario. The experiment yielded 89.52% average accuracy, improving the previous works of 77.36%.

## VI. CONCLUSION

In this work, we investigate the use of Indonesian and English text data with multilingual language models to improve Indonesian language's lack of data. Through experiments on sentiment analysis and hate speech detection task, we show that the addition of English text data with the utilization of multilingual language models can improve model performance on Indonesian tasks. The less the Indonesian language text data used, the greater improvement yielded from adding English text data. Finally, the full utilization of a pre-trained language model combined with added data by multilingual representation has successfully improved the result of Indonesian sentiment analysis and hate speech detection.

ACKNOWLEDGEMENT

This research is partly funded by ITB P3MI research for the Informatics research group in the School of Electrical Engineering and Informatics.

---

[3] https://suneeta-mall.github.io/2019/12/22/Reproducible-ml-tensorflow.html